\documentclass[10pt,twocolumn,letterpaper]{article}

\usepackage{cvpr}              

\usepackage{graphicx}
\usepackage{amsmath}
\usepackage{amssymb}
\usepackage{booktabs}
\usepackage{adjustbox}
\usepackage{makecell}
\usepackage{multirow}
\usepackage{bbding}

\usepackage[pagebackref,breaklinks,colorlinks]{hyperref}

\usepackage[capitalize]{cleveref}
\crefname{section}{Sec.}{Secs.}
\Crefname{section}{Section}{Sections}
\Crefname{table}{Table}{Tables}
\crefname{table}{Tab.}{Tabs.}

\newcommand\blfootnote[1]{%
  \begingroup
  \renewcommand\thefootnote{}\footnote{#1}%
  \addtocounter{footnote}{-1}%
  \endgroup
}

\def\cvprPaperID{5335} 
\def\confName{CVPR}
\def\confYear{2022}

\begin{document}

\title{PointCLIP: Point Cloud Understanding by CLIP}


\author{Renrui Zhang$^{*1}$, Ziyu Guo$^{*2}$, Wei Zhang$^{1}$, Kunchang Li$^{1}$, Xupeng Miao$^{2}$ \\ Bin Cui$^{2}$,  Yu Qiao$^1$, Peng Gao$^{ \dagger 1}$, Hongsheng Li$^{ \dagger 3}$\\ 
  $^1$Shanghai AI Laboratory \quad 
  $^2$Peking University\\
  $^3$The Chinese University of Hong Kong\\
\texttt{\{zhangrenrui, gaopeng, qiaoyu\}@pjlab.org.cn} \\
\texttt{2101210573@pku.edu.cn, hsli@ee.cuhk.edu.hk}
}

\maketitle
\blfootnote{$^*$ Equal contribution.\ \  $\dagger$ Corresponding author.}
\begin{abstract}

    Recently, zero-shot and few-shot learning via Contrastive Vision-Language Pre-training (CLIP) have shown inspirational performance on 2D visual recognition, which learns to match images with their corresponding texts in open-vocabulary settings. 
    However, it remains under explored that whether CLIP, pre-trained by large-scale image-text pairs in 2D, can be generalized to 3D recognition. In this paper, we identify such a setting is feasible by proposing \textbf{PointCLIP}, which conducts alignment between CLIP-encoded point cloud and 3D category texts. Specifically, we encode a point cloud by projecting it into multi-view depth maps without rendering, and aggregate the view-wise zero-shot prediction to achieve knowledge transfer from 2D to 3D. 
    On top of that, we design an inter-view adapter to better extract the global feature and adaptively fuse the few-shot knowledge learned from 3D into CLIP pre-trained in 2D. By just fine-tuning the lightweight adapter in the few-shot settings, the performance of PointCLIP could be largely improved. 
    In addition, we observe the complementary property between PointCLIP and classical 3D-supervised networks. By simple ensembling, PointCLIP boosts baseline's performance and even surpasses state-of-the-art models.
    Therefore, PointCLIP is a promising alternative for effective 3D point cloud understanding via CLIP under low resource cost and data regime. We conduct thorough experiments on widely-adopted ModelNet10, ModelNet40 and the challenging ScanObjectNN to demonstrate the effectiveness of PointCLIP. The code is released at \url{https://github.com/ZrrSkywalker/PointCLIP}.
   
\end{abstract}


\begin{figure}[t]
  \centering
\includegraphics[width=0.45\textwidth]{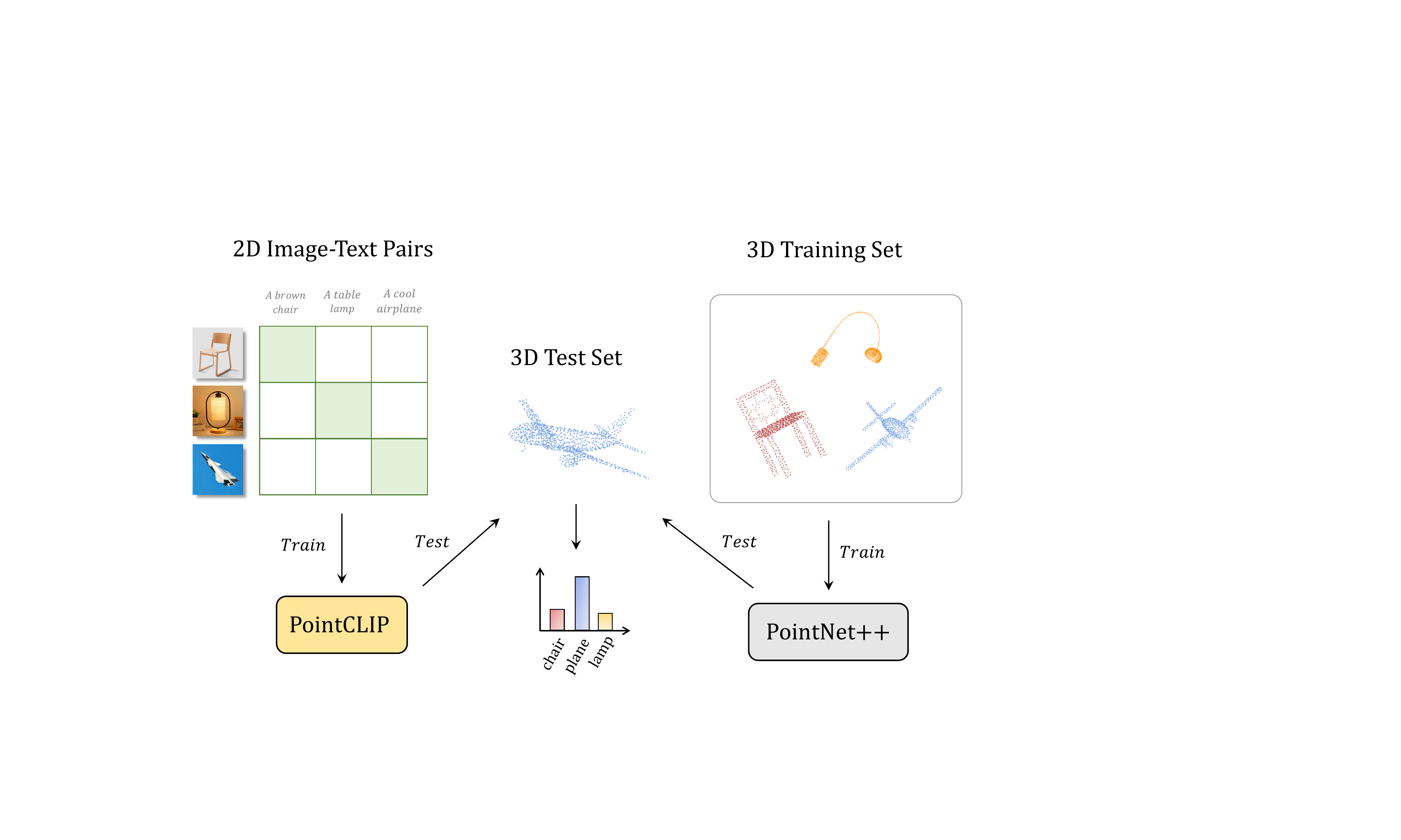}
   \caption{\textbf{A Comparison of Training-testing schemes between PointCLIP and PointNet++.} Different from classical 3D networks, our proposed PointCLIP is pre-trained by 2D image-text pairs, but conducts zero-shot classification on 3D datasets, which achieves cross-modality knowledge transfer.}
    \label{fig:simple_intro}
    \vspace{-0.5cm}
\end{figure}

\vspace{-15pt}
\section{Introduction}
\label{sec:intro}

Deep learning has dominated computer vision tasks of both 2D and 3D domains in recent years, such as image classification~\cite{he2016deep,dosovitskiy2021vit,mao2021dual,krizhevsky2012imagenet,parmar2018image,gao2021container}, object detection~\cite{ren2015faster,carion2020end,zheng2020end,lang2019pointpillars, engelcke2017vote3deep, chen2017multi}, semantic segmentation~\cite{zheng2021rethinking,zhang2018context,long2015fully,chen2017deeplab,huang2019ccnet}, point cloud recognition and part segmentation~\cite{qi2017pointnet,qi2017pointnet++,wang2019dynamic,goyal2021revisiting}. With 3D sensing technology developing rapidly, the growing demand for processing 3D point cloud data has boosted many advanced deep models with better local feature aggregator~\cite{li2018pointcnn,thomas2019kpconv,liu2019meteornet}, geometry modeling~\cite{guo2021pct,muzahid2020curvenet,pan20183dti} and projection-based processing~\cite{su2015multi,liu2019point,guo2021pct}. Different from grid-based 2D image data, 3D point clouds suffer from space sparsity and irregular distribution, which hinder direct methods transfer from 2D domain. Additionally, large-scale newly captured point cloud data contain a large number of objects of ``unseen" categories to the trained classifier. In this scenario, even the best-performing models might fail to recognize them and it is unaffordable to re-train every time when ``unseen" objects arise. 

Similar issues have been dramatically mitigated in 2D vision by Contrastive Vision-Language Pre-training (CLIP)~\cite{radford2021learning}, which proposed to learn transferable visual features with natural language supervisions. For zero-shot classification of ``unseen" categories, CLIP utilizes the pre-trained correlation between vision and language to conduct open-vocabulary recognition and achieves promising performance. To further enhance the accuracy in few-shot settings, CoOp~\cite{zhou2021coop} adopted learnable tokens to encode the text prompts, so that the classifier weights can be adaptively formed. From another perspective, CLIP-Adapter~\cite{gao2021clip} appends a lightweight residual-style adapter with two linear layers for better adapting image features. Tip-Adapter~\cite{zhang2021tip} further boosts its performance while greatly reduces the training time. Both methods achieve significant improvements over zero-shot CLIP. Consequently, the problem of recognizing new unlabeled objects has been explored by CLIP in 2D. However, a question is naturally arised: Could such CLIP-based models be transferred to 3D domain and realize zero-shot classification for ``unseen" 3D objects?

To address this issue, we propose \textbf{PointCLIP}, which transfers CLIP's 2D pre-trained knowledge to 3D point cloud understanding. The first concern is to bridge the modal gap between unordered point clouds and the grid-based images that CLIP could process. Considering the need for real-time prediction in various scenarios, such as autonomous driving~\cite{lang2019pointpillars, engelcke2017vote3deep, chen2017multi,qi2018frustum} and indoor navigation~\cite{zhu2017target}, we propose to adopt online perspective projection~\cite{goyal2021revisiting} without any post rendering~\cite{su2015multi}, i.e., simply projecting each point onto a series of pre-defined image planes to generate scatter depth maps. The cost of this projection process is marginal in both time and computation, but reserves the original property of the point cloud from multiple views. On top of that, we apply CLIP to encode multi-view features of point cloud by the CLIP pre-trained visual encoder and obtain each view's text-matched prediction independently via zero-shot classifier. Following CLIP, we place 3D category names into a hand-crafted template as prompts and generate the zero-shot classifier by CLIP's textual encoder. As different views contribute differently to the recognition of entire scene, we obtain the final prediction for point cloud by weighted aggregation between views.

Although PointCLIP achieves cross-modality zero-shot classification without any 3D training, its performance still falls behind classical point cloud networks well-trained on full datasets. To eliminate this gap, we introduce a learnable inter-view adapter with bottleneck linear layers to better extract features from multiple views in few-shot settings. Specifically, we concatenate all views' features and extract the compact global feature of the point cloud via interacting and summarizing cross-view information. Based on the global representation, adapted feature of each view is generated and added to their original CLIP-encoded feature via a residual connection. In this way, each view is equipped with the fused global feature and also combines newly adapted feature from the 3D few-shot dataset with 2D pre-trained CLIP's encoding. During training, we only fine-tune this lightweight adapter and freeze CLIP's both visual and textual encoders to avoid over-fitting, since only a few samples per class are given. Surprisingly, PointCLIP with an inter-view adapter with few-shot fine-tuning achieves comparable performance with some previous models well-trained with full datasets, which is a good trade-off between performance and cost.

Additionally, we observe that CLIP's 2D knowledge, supervised by contrastive loss, is complementary to the close-set 3D supervisions. The PointCLIP with an inter-view adapter can be fine-tuned under few-shot settings to improve the performance of classical fully-trained 3D networks. Taking PointCLIP in 16-shot ModelNet40~\cite{wu20153d} and fully-trained PointNet++~\cite{qi2017pointnet++} as an example, we directly ensemble their predicted logits for testing. Surprisingly, the performance of PointNet++'s 89.71$\%$, is enhanced to 92.03$\%$ by PointCLIP with an accuracy of 87.20$\%$. Furthermore, we select CurveNet~\cite{muzahid2020curvenet}, the state-of-the-art 3D recognition model, as the ensembling baseline, and achieve performance boost from 93.84$\%$ to 94.08$\%$. 
In contrast, simply ensembling two models fully trained on ModelNet40 without PointCLIP only leads to performance loss. Therefore, PointCLIP could be regraded as a multi-knowledge ensembling module, which promotes 3D networks via 2D contrastive knowledge with limited additional training.

The contributions of our paper are as follows:

\begin{itemize}
    \item We propose PointCLIP to extend CLIP for handling 3D point cloud data, which achieves cross-modality zero-shot recognition by transferring 2D pre-trained knowledge into 3D. 
    
    \item An inter-view adapter is introduced upon PointCLIP via feature interaction among multiple views and improves the performance of few-shot fine-tuning. 
    
    \item PointCLIP can be utilized as a multi-knowledge ensembling module for enhancing performance of existing fully-trained 3D networks, which surpasses state-of-the-art performances.
    
    \item Comprehensive experiments are conducted on widely adapted ModelNet10, ModelNet40 and the challenging ScanObjectNN, which indicate PointCLIP's potential for 3D understanding.
\end{itemize}

\section{Related Work}
\label{sec:related work}

\paragraph{Zero-shot Learning in 3D.}
The objective of zero-shot learning is to enable recognition of ``unseen" objects which are not adopted during training. Although zero-shot learning has drown much attention on 2D classification~\cite{radford2021learning,karessli2017gaze,xian2016latent}, only a few works explore how to conduct it in 3D domain. As the first attempt on point cloud, ~\cite{cheraghian2019zero} divides the 3D dataset into two parts: ``seen" and ``unseen" samples, and trains PointNet~\cite{qi2017pointnet} on the former but tests on the latter by measuring cosine similarities with category semantics. 
Based on this prior work, ~\cite{cheraghian2019mitigating} further mitigates the hubness problem~\cite{zhang2017learning} resulted from low-quality extracted 3D features and ~\cite{cheraghian2021zero} introduces a triplet loss for better performance in transductive settings, which allows to utilize unlabeled ``unseen" data at training time. 
Different from all above settings, which train the network on part of the 3D samples and predict on the others, PointCLIP achieves direct zero-shot recognition without any 3D training and conducts prediction on the whole point cloud datasets. Thus, our setting is more challenging for the domain gap between 2D pre-training and 3D application, but more urgent for practical problems.

\vspace*{-8pt}
\paragraph{Transfer Learning.}
Transfer learning~\cite{deng2009imagenet,zamir2018taskonomy} aims to utilize the knowledge from data-abundant domains to help the learning on data-scarce domains. For general vision, ImageNet~\cite{deng2009imagenet} pre-training can greatly assist downstream tasks, such as object detection~\cite{ren2015faster,carion2020end,gao2021fast} and semantic segmentation~\cite{long2015fully}. Also in natural language processing, representations pre-trained on web-crawled corpus via Mask Language Model~\cite{devlin2018bert} achieves leading performance on machine translation~\cite{mccann2017learned} and natural language inference~\cite{conneau2017supervised}.
Without any fine-tuning, the recently introduced CLIP~\cite{radford2021learning} shows superior image understanding ability for ``unseen" datasets. CLIP-Adapter~\cite{gao2021clip}, Tip-Adapter~\cite{zhang2021tip}, ActionCLIP~\cite{wang2021actionclip} and WiSE-FT~\cite{wortsman2021robust} further indicate that the performance of CLIP can be largely improved by infusing domain-specific supervisions. Although the successes stories are encouraging, most of the existing methods conduct knowledge transfer within the same modality, namely, image to image~\cite{deng2009imagenet}, video to video~\cite{carreira2017quo} or language to language~\cite{devlin2018bert}. Different from them, our PointCLIP is able to efficiently transfer representations learned from 2D images to the disparate 3D point clouds, which motivates future researches on transfer learning across different modalities.

\begin{figure*}[ht!]
  \centering
    \includegraphics[width=1\textwidth]{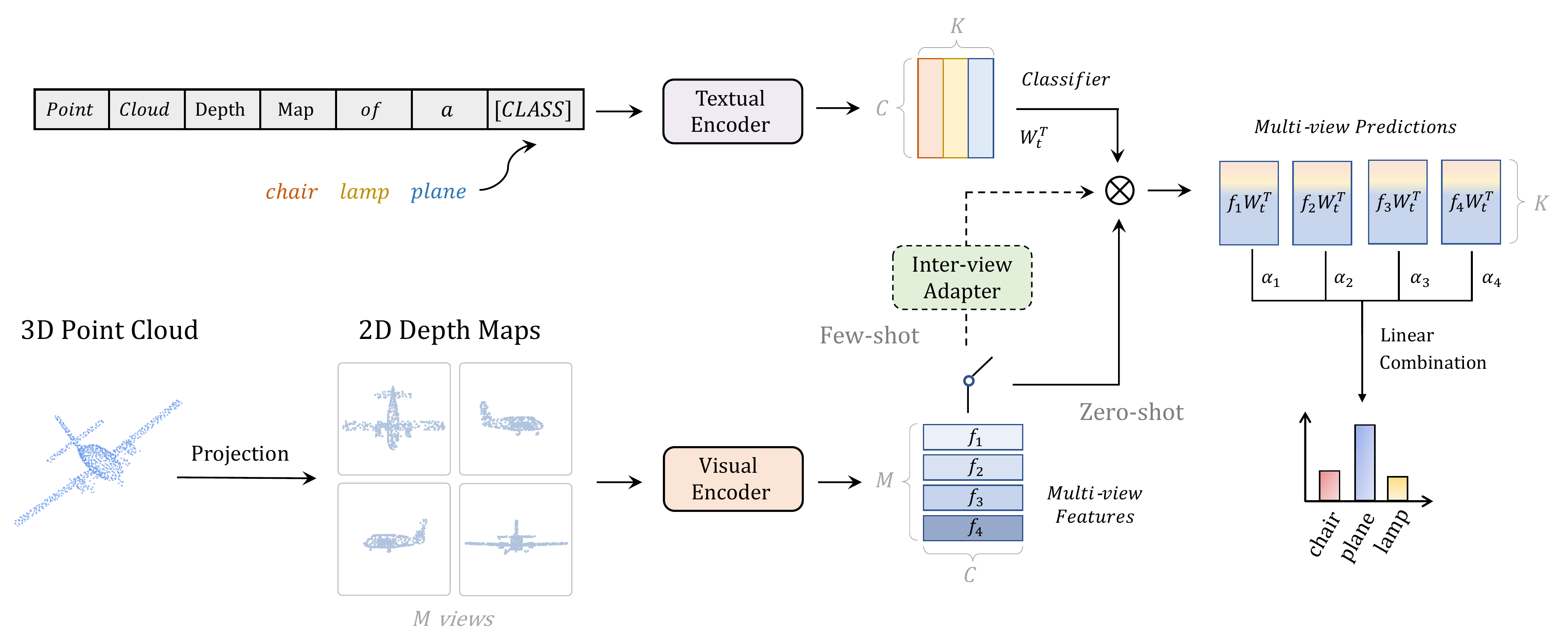}
   \caption{\textbf{The Pipeline of PointCLIP.} To bridge the modal gap, PointCLIP projects the point cloud onto multi-view depth maps, and conducts 3D recognition via CLIP pre-trained in 2D. The switch provides alternatives for direct zero-shot classification and few-shot classification with inter-view adapter, respectively, in solid and dotted lines.}
    \label{fig:cache_model}
    \vspace{-0.3cm}
\end{figure*}

\vspace*{-8pt}
\paragraph{Deep Neural Networks for Point Cloud.}
Existing deep neural networks for point cloud can be divided into point-based and projection-based methods. Point-based models process on raw points without any pre-transformation.
PointNet~\cite{qi2017pointnet} and PointNet++~\cite{qi2017pointnet++} firstly encode each point with a Multi-layer Perceptron (MLP) and utilize max pooling operation to realize permutation invariance. Recent point-based methods propose more advanced local aggregators and architecture designs~\cite{li2018pointcnn,thomas2019kpconv}. Other than raw points, projection-based methods understand point cloud by transferring it to volumetric~\cite{maturana2015voxnet} or multi-view~\cite{su2015multi} data forms. 
Therein, multi-view methods project point cloud into images of multiple views and process them with 2D Convolution Neural Networks (CNN)~\cite{he2016deep} pre-trained on ImageNet~\cite{krizhevsky2012imagenet}, such as MVCNN~\cite{su2015multi} and others~\cite{feng2018gvcnn, kanezaki2018rotationnet,yu2018mattnet,feng2019hypergraph}. Normally, such view-projected methods operate on offline-generated images which are projected from point-converted 3D meshes~\cite{wang20183d} or required post-rendering~\cite{sarkar2018learning} for shades and textures, so they are costly and impractical to be adopted for real-time applications. On the contrary, we follow SimpleView~\cite{goyal2021revisiting}, to naively project raw points onto image planes and set their pixel values according to the vertical distance. Such depth-map generation results in marginal time and computation costs, which meets the demand for efficient end-to-end zero-shot recognition.

\vspace*{-2pt}
\section{Method}
\label{sec:method}
In Section~\ref{reclip}, we first revisit Contrastive Vision-Language Pre-training (CLIP) for 2D zero-shot classification. Then in Section~\ref{pointclip}, we introduce our PointCLIP, which transfers 2D pre-trained knowledge into 3D. In Section~\ref{inter-view}, we provide PointCLIP with inter-view adapter for better performance under few-shot settings. In Section~\ref{m-ensemble}, we propose to ensemble PointCLIP with fully-trained classic 3D networks for multi-knowledge ensembling, which can achieve state-of-the-art performance.

\vspace*{1pt}
\subsection{A Revisit of CLIP}
\label{reclip}
CLIP is trained to match images with their corresponding natural language descriptions. There are two independent encoders in CLIP, respectively for visual and textual features encoding. During training, given a batch of images and texts, CLIP extracts their features and learns to align them in the embedding space with a contrastive loss. To ensure comprehensive learning, 400 million training image-text pairs are collected from the internet, which enables CLIP to align images with any semantic concepts in an open vocabulary for zero-shot classification.

Specifically, for an ``unseen" dataset of $K$ classes, CLIP constructs the textual inputs by placing all category names into a pre-defined template, known as prompt. Then, the zero-shot classifier, denoted as $W_t \in \mathbb{R}^{K \times C}$, is obtained by the $C$-dimensional textual feature of category prompts. Each of the $K$ row vectors in $W_t$ encodes the pre-trained category weights. Meanwhile, the feature of every test image is encoded by CLIP's visual encoder to $f_v \in \mathbb{R}^{1 \times C}$ and the classification $\mathrm{logits} \in \mathbb{R}^{1 \times K}$ are computed as,
\begin{align}
\label{clip}
\begin{split}
    \mathrm{logits} = f_v W_t^T; \ \ p_i = \mathrm{softmax}_i(\mathrm{logits}),
\end{split}
\end{align}
where $\mathrm{softmax}_i(\cdot)$ and $p_i$ denote the softmax function and predicted probability for category $i$. The whole process does not require new training images, but achieves promising zero-shot classification performance only by frozen pre-trained encoders.

\vspace*{1pt}
\subsection{Point Cloud Understanding by CLIP}
\label{pointclip}

A variety of large-scale datesets~\cite{krizhevsky2012imagenet, lin2014microsoft} in 2D provide abundant samples to pre-train models~\cite{he2016deep, dosovitskiy2020image} for high-quality and robust 2D features extraction. In contrast, the widely-adopted 3D datasets are comparatively much smaller and have limited categories, e.g. ModelNet40~\cite{wu20153d} with 9,843 samples and 40 classes vs. ImageNet~\cite{krizhevsky2012imagenet} with 1 million samples and 1,000 classes.
Thus, it is very difficult to obtain good pre-trained 3D networks for transfer learning. To alleviate this problem and explore the cross-modality power of CLIP, we propose PointCLIP to conduct zero-shot learning on point clouds based on the pre-trained CLIP.

\vspace*{-8pt}
\paragraph{Bridging the Modal Gap.}
Point cloud is a set of unordered points scattering in the 3D space, its sparsity and distribution greatly differ from grid-based 2D images. To convert point clouds into CLIP-accessible representations, we generate point-projected images from multiple views to eliminate the modal gap between 3D and 2D. In detail, if the coordinate of a point is denoted as $(x, y, z)$ in the 3D space, taking the bottom projection view as an example, its location on the image plane is $(\lceil x/z\rceil, \lceil y/z\rceil)$ following ~\cite{goyal2021revisiting}. In this way, the projected point cloud is a foreshortened figure, namely, small in the distance but big on the contrary, which is more similar to that in real photos. Other than ~\cite{goyal2021revisiting} applying convolution layers to pre-processing the one-channel depth map into three, we do not adopt any pre-convolution and directly set the pixel value equaling to $z$ in all three channels. Also, different from other off-line projection methods, whose projected images are generated from meshes~\cite{wang20183d} or CAD models~\cite{su2015multi}, our projected depth maps are from raw points and contain no color information but scattered depth values, which leads to marginal time and computation cost. With this lightweight cross-modality cohesion, CLIP's pre-trained knowledge can be then utilized for point cloud understanding.

\begin{figure*}[ht!]
\begin{minipage}[t]{0.63\textwidth}
\includegraphics[width=11cm]{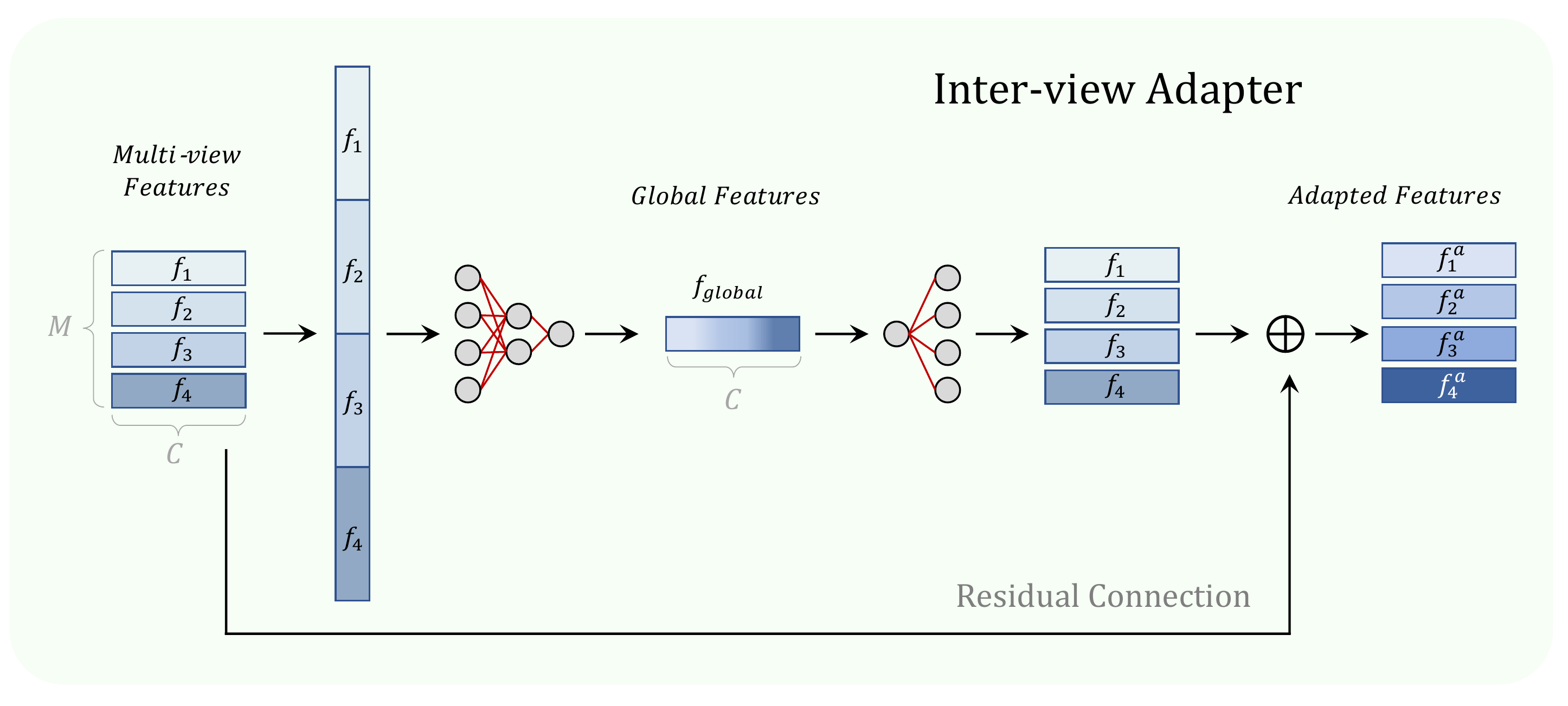}
\caption{Detailed structure of the proposed \textbf{Inter-view Adapter.} Given multi-view features of a point cloud, the adapter extracts its global representation and generates view-wise adapted features. Via a residual connection, the newly-learned 3D knowledge is fused into the pre-trained CLIP.}
\end{minipage}
\hspace{0.03in}
\begin{minipage}[t]{0.35\textwidth}

\includegraphics[width=6.1cm]{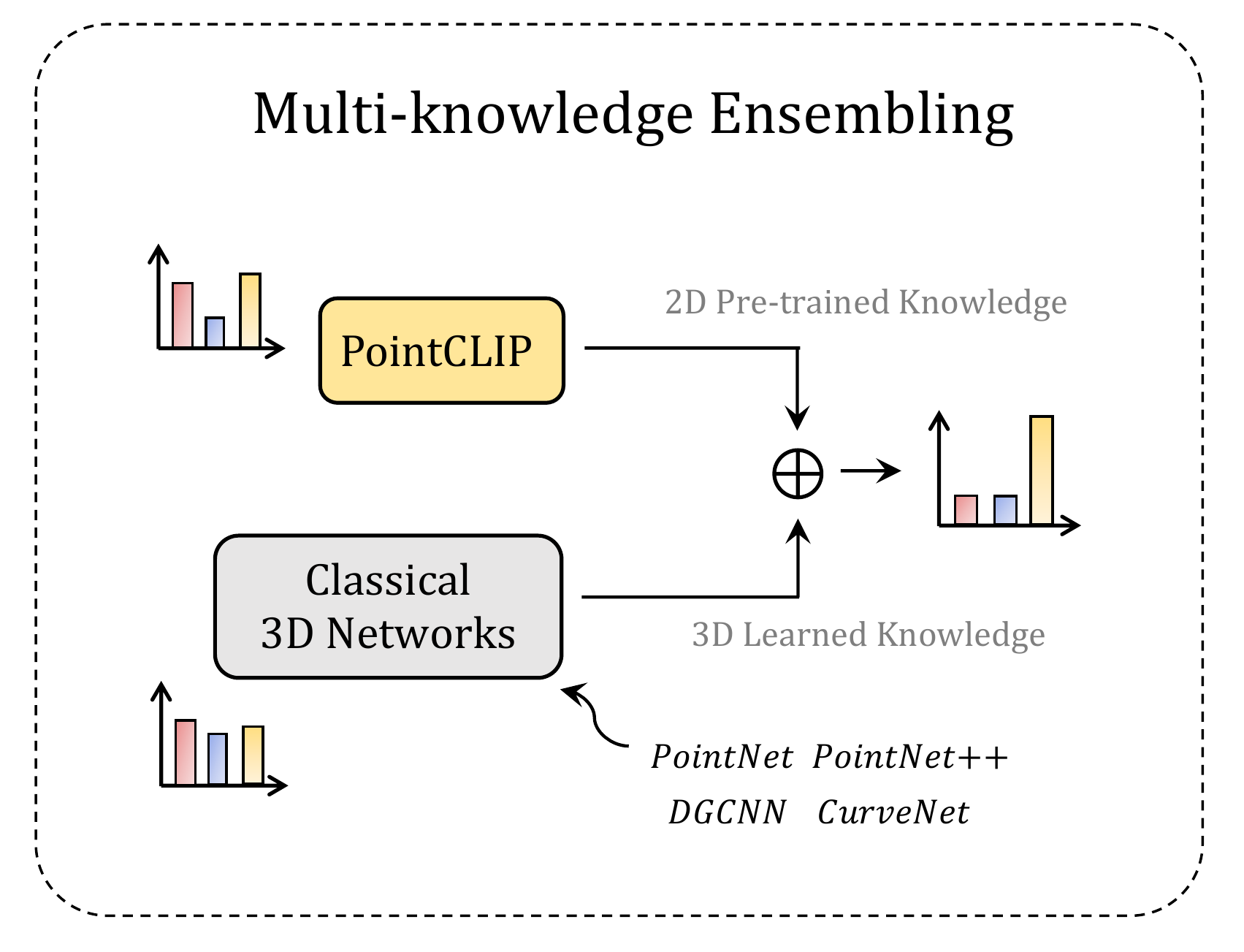}
\caption{PointCLIP could provide complimentary 2D knowledge to classical 3D networks and serve as a plug-and-play enhancement module.}
\end{minipage}
\vspace*{-15pt}
\end{figure*}

    
   

    
\vspace*{-12pt}
\paragraph{Zero-shot Classification.}
Based on projected images from $M$ views, we use CLIP to extract their visual features $\{f_i\}$, for $i=1,\dots,M$. For the textual branch, we place $K$ category names to the class token position of a pre-defined template: ``point cloud depth map of a [CLASS].'' and encode their textual features as the zero-shot classifier $W_t \in \mathbb{R}^{K \times C}$. On top of that, classification $\mathrm{logits}_i$ of each view are separately calculated and the final $\mathrm{logits}_p$ of point cloud are acquired by their weighted summation,
\begin{align}
\label{zeroshot}
\begin{split}
    &\mathrm{logits}_i = f_i W_t^T, \  \mathrm{for} \ \ i=1,\dots,M,\\
    &\mathrm{logits}_p = \sum\nolimits_{i=1}^M{\alpha_i \mathrm{logits}_i},
\end{split}
\end{align}
where $\alpha_i$ is a hyper-parameter weighing the importance of view $i$. Each view $f_i$ encodes a different perspective of the point cloud feature, which is capable for independent zero-shot classification. Their summation further complements the information of different perspectives to obtain an overall understanding. The whole process of PointCLIP is non-parametric for the ``unseen'' 3D dataset, which pairs each point cloud with its category via CLIP's pre-trained 2D knowledge and without any 3D training.

\vspace*{1pt}
\subsection{Inter-view Adapter for PointCLIP}
\label{inter-view}
Although PointCLIP achieves efficient zero-shot classification on point clouds, its performance is still incomparable to fully-trained 3D neural networks~\cite{qi2017pointnet,qi2017pointnet++}. We then consider a more common scenario where a few objects of each ``unseen'' category are contained in the newly collected data, and networks are required to recognize them under such few-shot settings. It is impractical to fine-tune the whole model, since the enormous parameters and insufficient samples would easily result in over-fitting. Therefore, referring to ~\cite{houlsby2019parameter} in Natural Language Processing (NLP) and CLIP-Adapter~\cite{gao2021clip} for fine-tuning pre-trained models on downstream tasks, we append a three-layer Multi-layer Perceptron (MLP) on top of PointCLIP, named inter-view adapter, to further enhance its performance under few-shot settings. For training, we freeze CLIP's both visual and textual encoders and fine-tune the learnable adapter via a cross-entropy loss.

To be specific, given CLIP-encoded $M$-view features of a point cloud, we concatenate them along the channel dimension as  $\operatorname{Concate}(f_{1\sim M}) \in \mathbb{R}^{1 \times MC}$, and acquire the compact global feature of point cloud via the first two layers of the inter-view adapter as
\begin{align}
\label{global}
\begin{split}
    f_\mathrm{global} = \operatorname{ReLU}(\operatorname{Concate}(f_{1\sim M}) W_1^T)W_2^T,
\end{split}
\end{align}
where $f_\mathrm{global} \in \mathbb{R}^{1 \times C}$ and $W_1$, $W_2$ stand for two-layer weights in the adapter. By this inter-view aggregation, features from multiple perspectives fuse into a summative representation. After that, the view-wise adapted feature is generated from the global feature and added to its original CLIP-encoded feature via a residual connection as
\begin{align}
\label{clip_}
\begin{split}
    f_i^a = f_i + \operatorname{ReLU}(f_\mathrm{global} 
    W_{3i}^T),
\end{split}
\end{align}
where $W_{3i} \in \mathbb{R}^{C \times C}$ denotes the $i$-th part of $W_3$ for view $i$, and $W_{3}^T = [W_{31}^T; W_{32}^T;\cdots W_{3M}^T]\in \mathbb{R}^{C \times MC}$. On the one hand, $f_i^a$ blends global-guided adapted feature into $f_i$ for the overall understanding of the point cloud and, thus, better view-wise prediction. On the other hand, the residual-style adapter infuses newly-learned 3D few-shot knowledge with that of 2D pre-trained CLIP, which further promotes the cross-modality knowledge transfer.

After the inter-view adapter, each view conducts classification with the adapted feature $f_i^a$ and the textual classifier $W_t$. Same as zero-shot classification, all $M$ logits from all views are summarized to construct the final prediction, and the view weights $\alpha_i$ can be learnable parameters here for more adaptive aggregation. Surprisingly, just fine-tuning this lightweight adapter with few-shot samples contributes to significant performance improvement, e.g. from 20.18$\%$ to 87.20$\%$ on ModelNet40 with 16 samples per category, less than 1/10 of the full data. This inspirational boost demonstrates the effectiveness and importance of feature adaption on 3D few-shot data, which greatly facilitates knowledge transfer from 2D to 3D. Consequently, PointCLIP with inter-view adapter provides a promising alternative solution for point cloud understanding. In some applications, there is no condition to train the entire model with large-scale fully annotated data, and fine-tuning only the three-layer adapter with few-shot data can achieve comparable performance.


\vspace*{1pt}
\subsection{Multi-knowledge Ensembling}
\label{m-ensemble}
Classical point cloud networks, such as the early PointNet~\cite{qi2017pointnet} and the recent CurveNet~\cite{muzahid2020curvenet}, are trained from scratch on 3D datasets by close-set supervision. In contrast, PointCLIP mostly inherits pre-trained priors from 2D vision-language learning, containing different aspects of knowledge. We then investigate if the two forms of knowledge can be ensembled together for joint inference. In practice, we first obtain the classical model, e.g. PointNet++~\cite{qi2017pointnet++} pre-trained from~\cite{simpleview2021}, and PointCLIP of either zero-shot or the adapter version. We conduct inferences of the two models and ensemble their predicted logits by simple addition as the final output. Beyond our expectation, aided by 16-shot fine-tuned PointCLIP of 87.20$\%$, PointNet++ of 89.71$\%$ is enhanced to 92.03$\%$ with a significant improvement of $+2.32\%$. In other words, ensembling of two low-score models can produce a much stronger one, which fully demonstrates the complimentary interaction of knowledge from the two models. Also, even with the zero-shot PointCLIP of 20.18$\%$, PointNet++ can still be improved to 92.10$\%$. In contrast, ensembling a pair of classical full-trained models would not enhance the performance, which indicates the importance of complimentary knowledge. We also implement this ensembling with other advanced networks and observe similar performance boosts, some of which achieve state-of-the-art performances. Therefore, PointCLIP can be utilized as a plug-and-play enhancement module to achieve robust point cloud understanding.

\begin{table}[t]
\centering
\begin{adjustbox}{width=\linewidth}
	\begin{tabular}{lccccc}
	\toprule
		\multicolumn{4}{c}{Zero-shot Performance of PointCLIP} \\ \midrule
		Datesets &Accuracy &Proj. Settings &View Weights \\
		\cmidrule(lr){1-1} \cmidrule(lr){2-2} \cmidrule(lr){3-3} \cmidrule(lr){4-4}
	    ModelNet10~\cite{wu20153d} &30.23\% &1.7, \ 100 & 2,5,7,10,5,6\\
	    ModelNet40~\cite{wu20153d}  &20.18\% &1.6, \ 121 &3,9,5,4,5,4\\
	    ScanObjectNN~\cite{uy2019revisiting} &15.38\% &1.8, \ 196 &3,10,7,4,1,0\\
	\bottomrule
	\end{tabular}
\end{adjustbox}
\caption{Zero-shot Performance of PointCLIP on ModelNet10, ModelNet40 and ScanObjectNN with the best-performing settings. Proj.Settings consist of projection distances and side length of the projected depth maps. View Weights are the relative values from 1 to 10.}
\vspace*{-7.4pt}
\label{zero-exp}
\end{table}

\begin{table}[t]
\centering
\begin{adjustbox}{width=\linewidth}
	\begin{tabular}{ccccccc}
	\toprule
	
        \multicolumn{7}{c}{View Numbers of Projection} \\
		\cmidrule(lr){1-7}
		Numbers  &1 &4 &\textbf{6} &8 &\textbf{10} &12\\ 
		
        \cmidrule(lr){1-1} \cmidrule(lr){2-7}
        \specialrule{0em}{1pt}{1pt}
		 Zero-shot &14.95  &18.68  &\textbf{20.18} &16.98 &14.91 &13.65\\ 
		 \specialrule{0em}{1pt}{1pt}
		 16-shot &75.53  &82.17  &84.24 &85.48 &\textbf{87.20} &86.35\\ 
		 \specialrule{0em}{1pt}{1pt}
      \midrule
      \midrule
        \multicolumn{7}{c}{Importance of each View} \\
		\cmidrule(lr){1-7}
        
	    View &Front &\textbf{Right} &Back &\textbf{Left} &Top &Down\\
	    
	     \cmidrule(lr){1-1} \cmidrule(lr){2-7}
	     \specialrule{0em}{1pt}{1pt}
	    Zero-shot &18.64  &\textbf{19.57} &18.92 &19.12 &17.46 &17.63 \\ 
	    \specialrule{0em}{1pt}{1pt}
	    16-shot &84.91  &85.69 &85.03 &\textbf{85.76} &84.44 &84.35 \\ 
	    \specialrule{0em}{1pt}{1pt}
	   
	\bottomrule
	\end{tabular}
\end{adjustbox}
\caption{Ablation studies ($\%$) concerning projected view numbers and each view's importance for zero-shot and 16-shot PointCLIP on ModelNet40.}
\vspace*{-5pt}
\label{ablation}
\end{table}

\section{Experiments}
\label{sec:experiments}

\vspace*{1pt}
\subsection{Zero-shot Classification}
\label{zero-exp_sec}

\begin{figure*}[ht]
    \centering
    
    \begin{minipage}[t]{0.33\linewidth}
    \centering
    \includegraphics[width=2.2in]{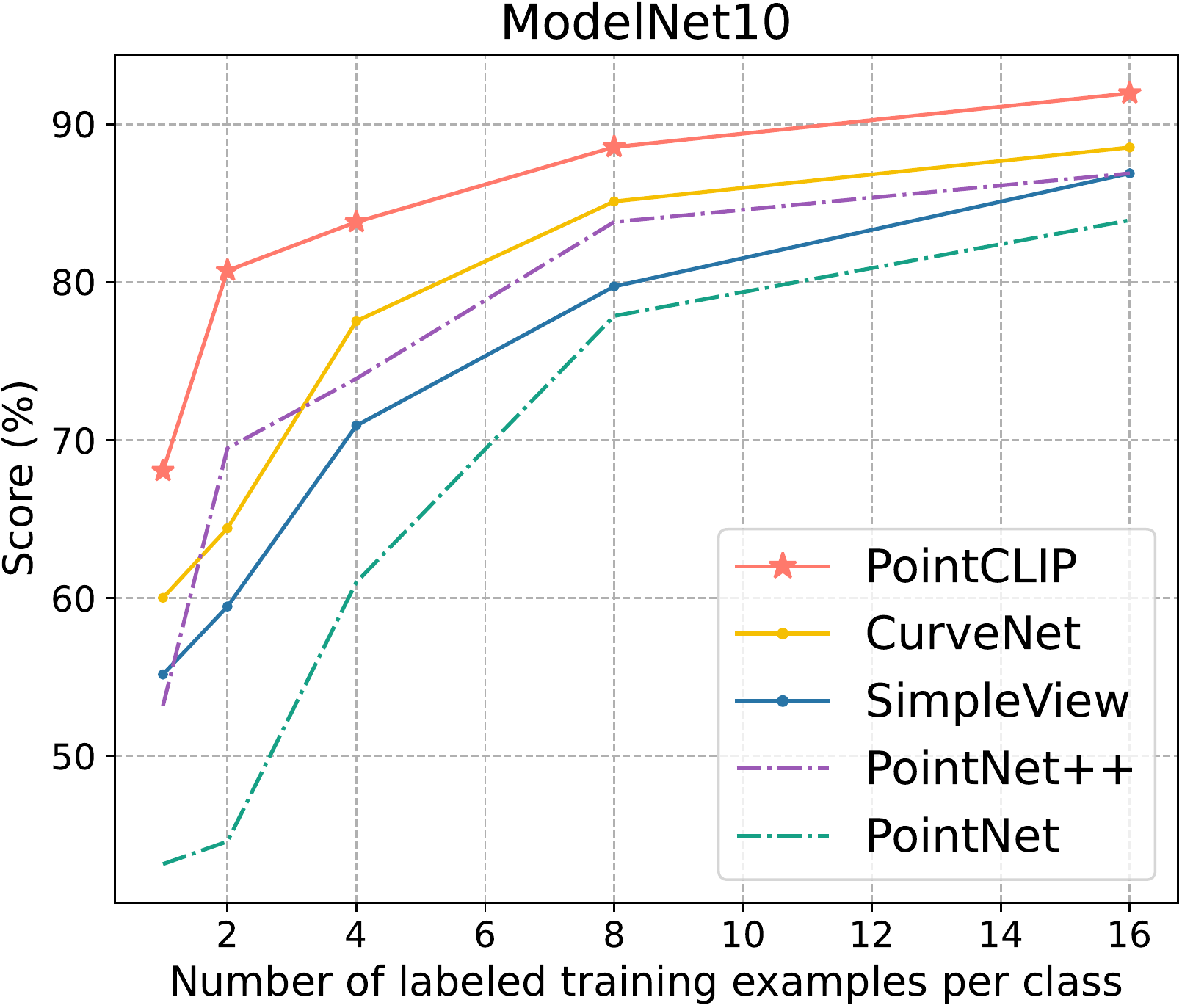}
    \end{minipage}
    \begin{minipage}[t]{0.33\linewidth}
    \centering
    \includegraphics[width=2.2in]{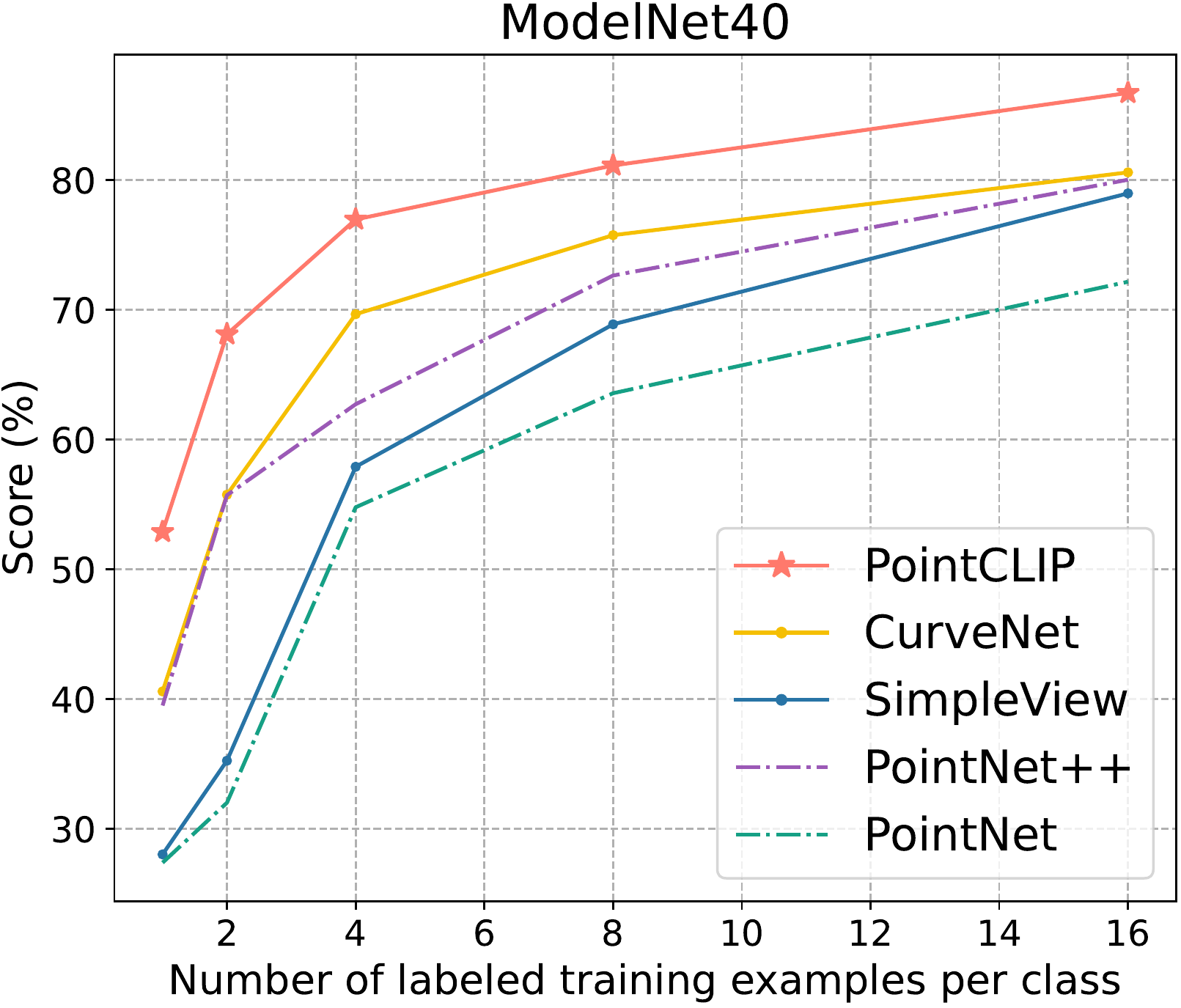}
    \end{minipage}
    \begin{minipage}[t]{0.33\linewidth}
    \centering
    \includegraphics[width=2.2in]{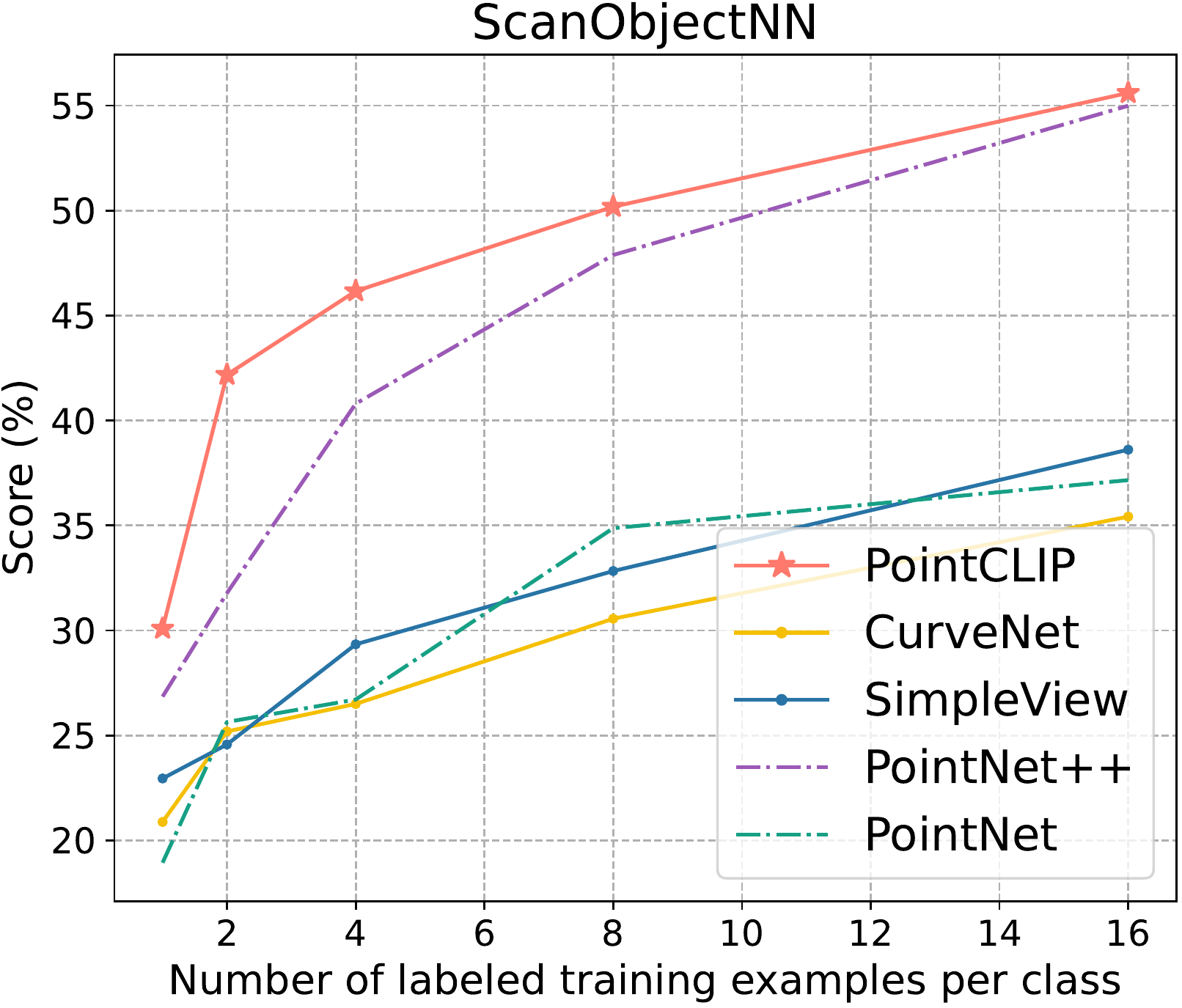}
    \end{minipage}

    \centering
    \caption{Few-shot performance comparison between PointCLIP and other classical 3D networks, including the state-of-the-art CurveNet, on ModelNet10, ModelNet40 and ScanObjectNN. Our PointCLIP shows consistent superiority to other models under 1, 2, 4, 8 and 16-shot settings.}
    \label{10 datasets}
    \vspace*{-12pt}
\end{figure*}

\paragraph{Settings.}
We evaluate the zero-shot classification performance of PointCLIP on three well-known datasets: ModelNet10~\cite{wu20153d}, ModelNet40~\cite{wu20153d} and ScanObjectNN~\cite{uy2019revisiting}. For each dataset, we require no training data and adopt the full test set for evaluation.
For the pre-trained CLIP model, we adopt ResNet-50~\cite{he2016deep} as the visual encoder and transformer~\cite{vaswani2017attention} as the textual encoder by default. We then project the point cloud from 6 orthogonal views: front, right, back, left, top and bottom, and each view has a relative weight value ranged from 1 to 10, shown in the fourth column of Table~\ref{zero-exp}. As the point coordinates are normalized from -1 to 1, we set the 6 image planes at a fixed distance away from the coordinate center $(0, 0)$. This distance is shown as the first value of Proj.Settings in Table~\ref{zero-exp}, and the larger distance leads to the denser points distributions on the image. The side length of projected square depth maps varies to different datasets, which is presented as the second value in Proj.Settings, and larger side length results in smaller projected object size. We then upsample all images to $(224, 224)$ for alignment with CLIP's settings. Also, we set the textual template as ``point cloud depth map of a [CLASS].'' to cater to the visual features of point clouds.

\begin{table}[t!]
\centering
\begin{adjustbox}{width=\linewidth}
	\begin{tabular}{lccc}
	\toprule
		Prompts & Zero-shot & 16-shot\\ \cmidrule(lr){1-1} \cmidrule(lr){2-2} \cmidrule(lr){3-3}
		``a photo of a [CLASS].'' &17.02\%&85.98\%\\
		``a point cloud photo of a [CLASS].'' &16.41\%&86.02\%\\
		``point cloud of a [CLASS].'' &18.68\% &86.06\%\\
		``point cloud of a big [CLASS].'' &19.21\%&\textbf{87.20}\%\\
		``point cloud depth map of a [CLASS].'' &\textbf{20.18}\% &85.82\%\\
		``[Learnable Tokens] + [CLASS]'' &- &73.63\%\\
		
	\bottomrule
	\end{tabular}
\end{adjustbox}
\caption{Performances of PointCLIP with different prompt designs on ModelNet40. [CLASS] denotes the class token, and [Learnable Tokens] denotes learnable prompts with fixed length.}
\vspace*{-5pt}
\label{prompt}
\end{table}

\begin{table}[t]
\centering
\vspace*{-0.3pt}
\begin{adjustbox}{width=\linewidth}
	\begin{tabular}{lccccccc}
	\toprule
		\multicolumn{7}{c}{Different Visual Encoders} \\
		\midrule
		Models &RN50 &\textbf{RN101} &ViT/32 &ViT/16 &RN.$\times$4 &\textbf{RN.$\times$16}\\
        \cmidrule(lr){1-1} \cmidrule(lr){2-7}
        \specialrule{0em}{1pt}{1pt}
		 Zero-shot &20.18 &17.02  &16.94  &21.31 &17.02 &\textbf{23.78}\\ 
		 \specialrule{0em}{1pt}{1pt}
        16-shot &85.09 &\textbf{87.20}  &83.83  &85.37 &85.58 &85.90\\ 
		 \specialrule{0em}{1pt}{1pt}
	   
	\bottomrule
	\end{tabular}
\end{adjustbox}
\caption{Performances (\%) of PointCLIP for different visual encoders on ModelNet40. RN50 denotes ResNet-50, and ViT-B/32 represents vision transformer with 32 $\times$ 32 patch embeddings, and RN.$\times$16 denotes ResNet-50 with 16 times more computations from~\cite{radford2021learning}.}
\vspace*{-8pt}
\label{encoder}
\end{table}

\vspace*{-5pt}
\paragraph{Performance.} In Table~\ref{zero-exp}, we present performances of zero-shot PointCLIP for three datasets with their best-performing settings. Without any 3D training, PointCLIP is able to achieve a promising 30.23$\%$ on ModelNet10, which demonstrates the effective knowledge transfer from 2D to 3D. For ModelNet40 with 4 times the number of categories and ScanObjectNN of noisy real-world scenes, PointCLIP achieves slightly worse performances, 20.18$\%$ and 15.38$\%$, respectively, due to the lack of 3D downstream adaptions. As for the projection distances and image resolutions of Proj.Settings, their variances accord with the properties of different datasets. Compared to indoor ModelNet10, PointCLIP on ModelNet40 requires more details to recognize complex outdoor objects, such as airplanes and plants, and thus performs better with more scattered points and larger object size, namely, larger perspective projection distance and resolutions. In contrast, for ScanObjectNN, denser points and larger resolutions are required for filtering out the noise and reserving complex real-scene information. With respect to view weights, ModelNet10 and ModelNet40 of synthetic objects require all 6 views' contributions to the final classification with different importance, but for ScanObjectNN which contains noisy points of floors and ceilings, the top and bottom views could hardly provide any information.

\vspace*{-8pt}
\paragraph{Ablations.}
\label{abzero}
In Table~\ref{ablation}, We conduct ablation studies of zero-shot PointCLIP concerning projection view numbers and the importance of each view on ModelNet40. For the number of projected views, we try 1, 4, 6, 8, 10 and 12\footnote{The settings of views are in the Appendix.} views, for increasingly capturing the multi-view information of point clouds, but more than 6 views would bring redundancy and lead to performance decay. To explore how different views impact the performance, we unify all relative weights to 3 and respectively increase each view's weight to 9. As is shown in the table, projection from the right achieves the highest performance, which indicates its leading role, and the top and down views contribute relatively less to the zero-shot classification. In Table~\ref{encoder}, we implement different visual backbones from ResNet~\cite{he2016deep} to vision transformer~\cite{dosovitskiy2020image}, and RN50$\times$16~\cite{radford2021learning} achieves the best performance of 23.78$\%$, which has 16 times more computations than ResNet-50. However, upgrading ResNet-50 to ResNet-101 with more parameters and deeper layers would not provide higher classification accuracy.

\vspace*{-8pt}
\paragraph{Prompt Design.} We present five prompt designs for zero-shot PointCLIP in Table~\ref{prompt}. We observe that the naive ``a photo of a [CLASS].'' achieves 17.02$\%$ on ModelNet40, but simply inserting the word ``point cloud'' into it would hurt the performance. We then remove ``a photo'' and directly utilize ``point cloud'' as the subject, which benefits the accuracy by +1.66$\%$. Also, as the projected point cloud normally covers most of the image area, appending an adjective ``big'' could bring further performance improvement. Furthermore, we add the ``depth map'' to describe the projected images more relevantly, which contributes to the best-performing 20.18$\%$, demonstrating the importance of prompt choices.
\label{Prompt Design}

\vspace*{1pt}
\subsection{Few-shot Classification}
\paragraph{Settings.} We experiment PointCLIP with the inter-view adapter under 1, 2, 4, 8, 16 shots also in the three datasets: ModelNet10~\cite{wu20153d}, ModelNet40~\cite{wu20153d} and ScanObjectNN~\cite{uy2019revisiting}. For $K$-shot settings, we randomly sample $K$ point clouds from each category of the training set. We inherit the best projection settings from zero-shot experiments in Section~\ref{zero-exp_sec}. In contrast, considering both efficiency and performance, we adopt ResNet-101~\cite{he2016deep} as CLIP's pre-trained visual encoder for stronger feature extraction, and increase the projected view numbers to 10, adding the views of upper/bottom-front/back-left corners, since the left view is proven to be the most informative for few-shot recognition in Table~\ref{ablation}. In addition, we modify the prompt to ``point cloud of a big [CLASS].'', which performs better in the few-shot experiments. For the inter-view adapter, we construct a residual-style Multi-layer Perceptron (MLP) consisting of three linear layers, as described in Section~\ref{inter-view}.


\vspace*{-8pt}
\paragraph{Performance.} In Figure~\ref{10 datasets}, we present the few-shot performances of PointCLIP and compare it with 4 representative 3D networks: PointNet~\cite{qi2017pointnet}, PointNet++~\cite{qi2017pointnet++}, SimpleView~\cite{goyal2021revisiting} and the state-of-the-art CurveNet~\cite{muzahid2020curvenet}. As we can see, PointCLIP with inter-view adapter surpasses all other methods for the few-shot classification. When there are only a small number of samples per category, PointCLIP has distinct advantages, exceeding PointNet by 25.49$\%$ and CurveNet by 12.29$\%$ on ModelNet40 with 1 shot. When given more training samples, PointCLIP still leads the performance, but the gap becomes smaller due to the limited fitting capacity of the lightweight three-layer adapter. For the detailed training settings, please refer to the Appendix.

\vspace*{-8pt}
\paragraph{Ablations.}
\label{abfew}
In Table~\ref{ablation}, we show the 16-shot PointCLIP under different projection views and explore how each view contributes on ModelNet40. Differing from the zero-shot version, 10 views of 16-shot PointCLIP performs better than 6 views, probably because the newly-added adapter is able to better utilize the information from more views and adaptively aggregate them. For the importance of views, we follow the configurations of our zero-shot version and observe the reversed conclusion that, the left view is the most informative here. Surprisingly, for different visual encoders in Table~\ref{encoder}, ResNet-101 achieves the highest accuracy with less parameters than vision transformer or ResNet-50$\times$16. Table~\ref{prompt} lists the performance influence caused by prompt designs, and the ``point cloud of a big [CLASS].'' performs the best, which is slightly different from the analysis in Paragraph~\ref{Prompt Design}.

\begin{table}[t]
\centering
\vspace*{13pt}
\begin{adjustbox}{width=\linewidth}
	\begin{tabular}{lcccccc}
	\toprule
		Models & Before En. & After En. & Gain &Ratio\\ \midrule
		PointNet~\cite{qi2017pointnet} &88.78 &90.76 &\color{blue}{+1.98} &0.60\\
		PointNet++~\cite{qi2017pointnet++} & 89.71 & 92.10 &\color{blue}{+2.39} &0.70\\
		RSCNN~\cite{liu2019relation} & 92.22 & 92.59 &\color{blue}{+0.37} &0.70\\
		DGCNN~\cite{wang2019dynamic} & 92.63 & 92.83  &\color{blue}{+0.20} &0.70\\
		SimpleView~\cite{goyal2021revisiting} & 93.23 & 93.87 & \color{blue}{+0.64} &0.60\\
		CurveNet~\cite{muzahid2020curvenet} & 93.84 & \textbf{94.08} & \color{blue}{+0.24} &0.15\\
	\bottomrule
	\end{tabular}
\end{adjustbox}
\caption{The enhancement ability ($\%$) of 16-shot PointCLIP, which achieves 87.20$\%$, on multiple classical 3D networks in ModelNet40. Before and After En. denote models with and without PointCLIP ensembling, respectively.}
\vspace*{-12pt}
\label{ensemble}
\end{table}

\vspace*{1pt}
\subsection{Multi-knowledge Ensembling}

\paragraph{Settings.} To verify the complementarity of blending pre-trained 2D priors with 3D knowledge, we aggregate the fine-tuned 16-shot PointCLIP of 87.20$\%$ on ModelNet40, respectively with fully-trained PointNet~\cite{qi2017pointnet}, PointNet++~\cite{qi2017pointnet++}, DGCNN~\cite{wang2019dynamic}, SimpleView~\cite{goyal2021revisiting} and CurveNet~\cite{muzahid2020curvenet}, whose trained models are obtained from ~\cite{simpleview2021,curvenet} without any voting. We manually modulate the fusion ratio between PointCLIP and each model, and report the performance with the best Ratio in Table~\ref{ensemble}, which represents PointCLIP's relative weight to the whole.

\vspace*{-8pt}
\paragraph{Performance.} As shown in Table~\ref{ensemble}, ensembling with PointCLIP improves the performances of all classical fully-trained 3D networks. The results fully demonstrate the complementarity of PointCLIP to existing fully-trained 3D models, and the performance gain is not simply achieved by ensembling models. These are surprising results to us, because the accuracy of 16-shot PointCLIP is lower than all other models trained with full datasets, but could still benefit their already high performances to be higher. Therein, the largest accuracy improvement is on PointNet++ from 89.71$\%$ to 92.10$\%$, and combining PointCLIP with the state-of-the-art CurveNet further achieves 94.08$\%$. Also, we observe that, for models with low baseline performances, PointCLIP's logits need to account for a large proportion, but for the well-performing ones, such as CurveNet, their knowledge is supposed to play a dominant role in the ensembling. 

\vspace*{-8pt}
\paragraph{Ablations.} We conduct ablation studies of ensembling two models fully trained on ModelNet40 without PointCLIP, and fuse their logits with the same ratio for simplicity. As is presented in Table~\ref{ensembleab}, ensembling PointNet++ lowers the performance of RSCNN and CurveNet, and aggregating the highest two models, SimpleView and CurveNet, could not achieve better performance. Also, a pair of PointCLIP would hurt the performance. Hence, simply ensembling two models with the same training scheme normally leads to performance degradation, which demonstrates the significance of multi-knowledge interaction. In Table~\ref{128}, we fuse zero-shot PointCLIP and the model fine-tuned by 8, 16, 32, 64 and 128 shots, respectively with CurveNet to explore their ensembling performances. As reported, zero-shot PointCLIP with only 20.18$\%$ could enhance CurveNet by +0.04$\%$. However, too much training on 3D dataset would adversely influence the ensembling accuracy. This is possibly caused by the too high similarity between two models, which cannot provide complementary knowledge as expected.

\begin{table}[t]
\centering
\vspace*{13pt}
\begin{adjustbox}{width=\linewidth}
	\begin{tabular}{lcccc}
	\toprule
		En. Model 1 & &En. Model 2 & After En.\\ \midrule
		PointNet++~\cite{qi2017pointnet++}, 89.71 &+ &RSCNN~\cite{liu2019relation}, 92.22 &92.14\\
		PointNet++, 89.71 &+ &CurveNet~\cite{muzahid2020curvenet}, 93.84 &91.61\\
		SimpleView~\cite{goyal2021revisiting}, 93.23 &+ &CurveNet, 93.84 &93.68\\
		PointCLIP, 87.20  &+ &PointCLIP, 87.14 &87.06\\
	\bottomrule
	\end{tabular}
\end{adjustbox}
\caption{Ablation studies ($\%$) of ensembling models both trained on ModelNet40 or pre-trained in 2D.}
\vspace*{-12pt}
\label{ensembleab}
\end{table}

\begin{table}[t]
\centering
\vspace*{-0.3pt}
\begin{adjustbox}{width=\linewidth}
	\begin{tabular}{lccccccc}
	\toprule
		\multicolumn{7}{c}{Ensembling with CurveNet~\cite{muzahid2020curvenet}} \\
		\midrule
		Shots &0 &8 &16 &32 &64 &128\\
        \cmidrule(lr){1-1} \cmidrule(lr){2-7}
        \specialrule{0em}{1pt}{1pt}
		 PointCLIP &20.18 &81.96   &87.20   &87.83   &88.95   &\textbf{90.02}\\ 
		 \specialrule{0em}{1pt}{1pt}
        After En. &93.88  &93.89  &\textbf{94.08}   &94.00   &93.92   &93.88\\ 
		 \specialrule{0em}{1pt}{1pt}
	   
	\bottomrule
	\end{tabular}
\end{adjustbox}
\caption{Enhancement performance ($\%$) of PointCLIP under different few-shot settings for CurveNet on ModelNet40.}
\vspace*{-12pt}
\label{128}
\end{table}

\section{Conclusion and Limitation}
\label{sec:conclusion}

We propose PointCLIP, which conducts cross-modality zero-shot recognition on point cloud without any 3D training. Via multi-view projection, PointCLIP efficiently transfers CLIP's pre-trained 2D knowledge into the 3D domain. Under few-shot settings, we design a lightweight inter-view adapter to aggregate multi-view representations and generate adapted features. By fine-tuning such adapter and freezing all other modules, the performance of PointCLIP is largely improved. In addition, PointCLIP could serve as a plug-and-play module to provide complimentary information for the classical 3D networks, which surpasses state-of-the-art performance. Although PointCLIP realizes the transfer learning from 2D to 3D, how to utilize CLIP's knowledge for other 3D tasks is still under explored. Our future work will focus on generalizing CLIP for wider 3D applications.

{\small
\bibliographystyle{ieee_fullname}
\bibliography{egbib}
}

\clearpage
\appendix

\section*{Appendix}
\maketitle

\section{Datasets}
We evaluate our PointCLIP on three well-known datasets: ModelNet10~\cite{wu20153d}, ModelNet40~\cite{wu20153d} and ScanObjectNN~\cite{uy2019revisiting}. Therein, ModelNet10 consists of 4,899 synthetic meshed CAD models with 10 indoor categories, 3,991 for training and 908 for testing. ModelNet40 is larger and contains 12,311 samples of 40 common categories, 9,843 for training and 2,468 for testing. In both datasets, we uniformly sample 1,024 points from each object as the network input. ScanObjectNN contains 2,321 training and 581 testing point clouds of 15 categories collected directly from real-world scans. Different from synthetic data with complete profiles, objects in ScanObjectNN are occluded at different levels and disturbed with background noise, so it is more challenging for accurate recognition.

\section{Implementation Details}
For ablation studies of projected view numbers, we adopt different settings for zero-shot and few-shot PointCLIP. As the right view is the most important for zero-shot PointCLIP, we set the 12 views to: front, right, back, left, top, bottom, upper/lower right diagonal front/back (4 views) and upper left diagonal front/back (2 views). In contrast, few-shot PointCLIP achieves higher performance with left views, so we replace all the ``left'' settings above into ``right''. For both versions, the view number of $M$ represents picking the first $M$ views for experiments.

For PointCLIP with inter-view adapter, we fine-tune it under 1, 2, 4, 8 and 16 shots with batch size 32 and learning rate 0.01 for 250 epochs. Stochastic Gradient Decent (SGD)~\cite{kingma2014adam} with momentum 0.9 is adopted as the optimizer. We utilize a cosine scheduler for learning rate decay and Smooth Loss~\cite{wang2019dynamic} following ~\cite{goyal2021revisiting}. In ModelNet10 and ModelNet40, We apply random scaling and translation for training augmentation, but in the challenging ScanObjectNN, we append jitter and random rotation following~\cite{qi2017pointnet}. During training, we freeze CLIP's both visual and textual encoders, and only fine-tune the inter-view adapter. For other compared models, we unfreeze all the parameters, and adopt the same data augmentation and loss functions reported in the papers.
\section{Supplementary Ablations}
\paragraph{Inter-view Adapter.}
We adopt the inter-view adapter with three linear layers: one for global extraction and two for view-wise adapted features generation. Here, we explore other architectures of the adapter on 16-shot PointCLIP for ModelNet40 in Table~\ref{adapter}. Specifically, w/o global denotes the adapter processing each view separately without interaction, and the w/o view-wise version repeats the global feature as each view's adapted feature. The 2-layer adapter removes the linear layer after the global representation and the pre-layer version moves it before the global extraction. The results show that dropping or changing the original modules in the adapter would all hurt the performance, especially the inter-view extraction of global feature.

\begin{table}[h!]
\centering
\vspace*{-0.3pt}
\begin{adjustbox}{width=\linewidth}
	\begin{tabular}{lccccc}
	\toprule
		\multicolumn{5}{c}{Architectures of Inter-view Adapter} \\
		\midrule
		original &w/o global &w/o view-wise &2-layer &pre-layer\\
        \cmidrule(lr){1-1} \cmidrule(lr){2-5}
        \specialrule{0em}{1pt}{1pt}
		 87.20 &83.87 &85.93  &86.48   &86.78   \\ 
		 \specialrule{0em}{1pt}{1pt}
	   
	\bottomrule
	\end{tabular}
\end{adjustbox}
\caption{Architectures of the inter-view adapter.}
\vspace*{-20pt}
\label{adapter}
\end{table}

\paragraph{Adapted Features Fusion.}
The view-wise adapted feature is generated by the adapter and then added to the original CLIP-encoded feature via a residual connection. On ModelNet40, we evaluate the performance of 16-shot PointCLIP with different fusion ratios $\beta$, which denotes the proportion of adapted features. To show the effect of $\beta$, we set all view weights the same. From the results in Table~\ref{ratio}, different ratios actually lead to little performance variance and the $\beta$ of 0.6 perfoms better than others. Thus, we adopt 0.6 as the fusion ratio by default, which indicates the comparable contributions between 2D pre-trained knowledge and 3D learned knowledge.
\begin{table}[ht]
\centering
\begin{adjustbox}{width=\linewidth}
	\begin{tabular}{lccccccccccc}
	\toprule
		\multicolumn{11}{c}{Adapter Fusion Ratios $\beta$} \\
		\midrule
		0.0 &0.1 &0.2 &0.3 &0.4 &0.5 &0.6 &0.7 &0.8 &0.9 &1.0\\
        \cmidrule(lr){1-11}
        \specialrule{0em}{1pt}{1pt}
		 9.56 &85.74 &85.78 &85.66 &85.76 &85.98 &86.13 &85.91 &85.85 &85.74 &85.53  \\ 
		 \specialrule{0em}{1pt}{1pt}
	   
	\bottomrule
	\end{tabular}
\end{adjustbox}
\caption{Different fusion ratios of adapted features.}
\vspace*{-24pt}
\label{ratio}
\end{table}

\begin{figure*}[ht]
  \centering
\includegraphics[width=0.8\textwidth]{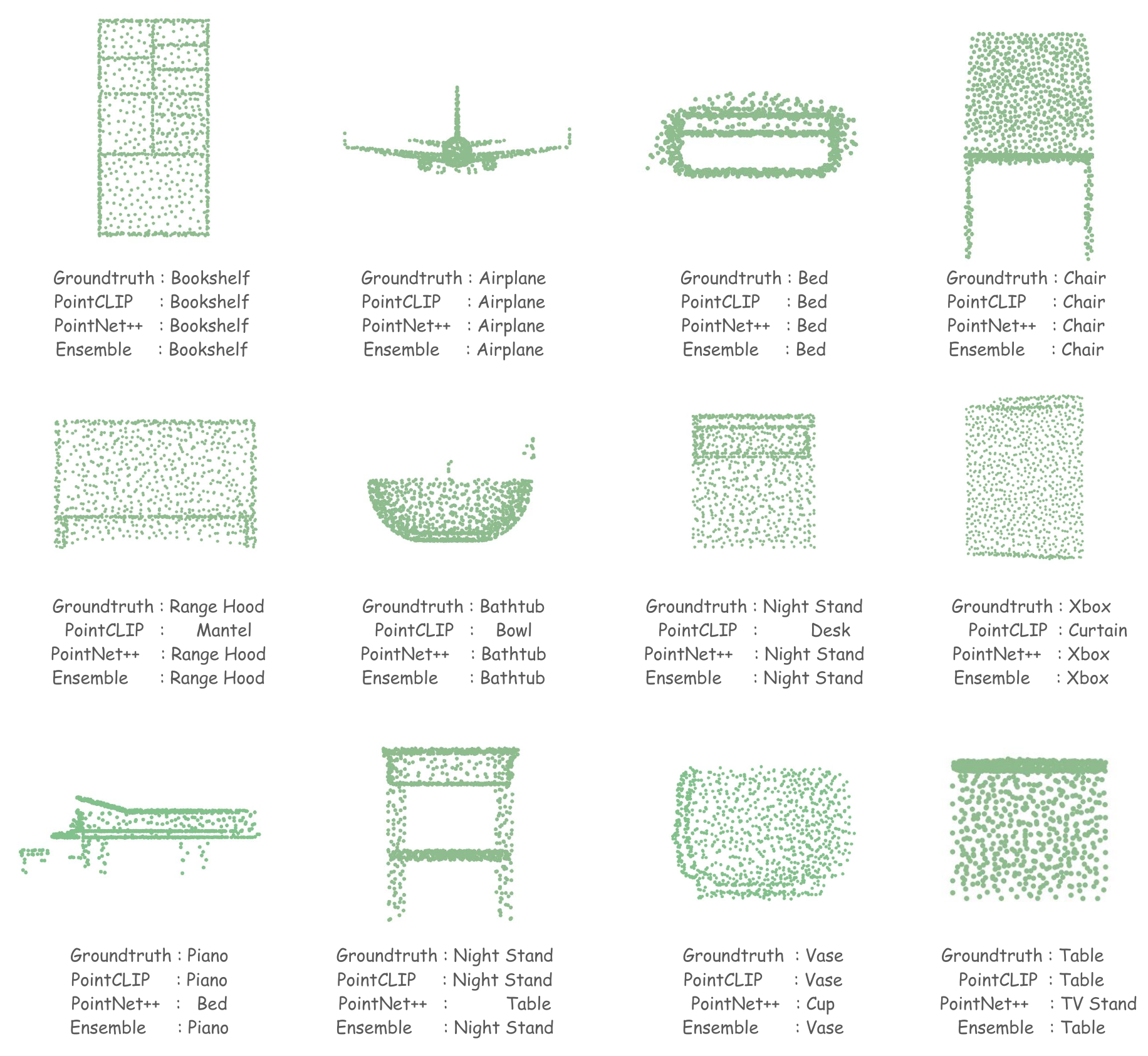}
   \caption{Visualizations of predictions by PointCLIP, PointNet++~\cite{qi2017pointnet++} and the ensembled model.}
    \label{fig:ensemble}
    \vspace{0cm}
\end{figure*}

\paragraph{Full Training Set.}
We also fine-tune PointCLIP on full training set of ModelNet40~\cite{wu20153d} and present the results in Table~\ref{encoder_}. Likewise, we freeze both pre-trained visual and textual encoders in CLIP and only train the inter-view adapter. As expected, visual encoders with more parameters lead to higher accuracy, and only fine-tuning the appended lightweight adapter could achieve the performance of 92.01$\%$.
\begin{table}[h!]
\centering
\vspace*{-0.3pt}
\begin{adjustbox}{width=\linewidth}
	\begin{tabular}{lccccccc}
	\toprule
		\multicolumn{7}{c}{Fine-tuning on Full ModelNet40~\cite{wu20153d}} \\
		\midrule
		Models &RN50 &RN101 &ViT/32 &ViT/16 &RN.$\times$4 &\textbf{RN.$\times$16}\\
        \cmidrule(lr){1-1} \cmidrule(lr){2-7}
        \specialrule{0em}{1pt}{1pt}
		 Accuracy &86.42 &91.69  &91.76  &90.70 &91.93 &\textbf{92.01}\\ 
		 \specialrule{0em}{1pt}{1pt}
	   
	\bottomrule
	\end{tabular}
\end{adjustbox}
\caption{Performances (\%) for fine-tuning PointCLIP on full training set of ModelNet40 with different visual encoders.}
\vspace*{-2pt}
\label{encoder_}
\end{table}

\paragraph{Fine-tuning Settings.}
Under full training set of ModelNet40~\cite{wu20153d}, we further fine-tune different modules of PointCLIP in Table~\ref{fine-tune}. Therein, we adopt ResNet-101~\cite{he2016deep} as the visual encoder, and fine-tuning without the adapter represent unfreezing the visual or textual encoder upon the zero-shot PointCLIP. As presented, unfreezing just the textual encoder normally hurts the performance, but training both encoders and all modules of PointCLIP achieves better performance of 91.40$\%$ and 91.89$\%$, respectively.
\begin{table}[ht]
\centering
\begin{adjustbox}{width=\linewidth}
	\begin{tabular}{cccc}
	\toprule
		 Visual Encoder & Textual Encoder 
		 & Inter-view Adapter  & Accuracy($\%$)\\ \midrule
		\Checkmark&- &- &91.01 \\
		- &\Checkmark &- &73.89 \\
		\Checkmark & \Checkmark & - & 91.49 \\
		- & - & \Checkmark & 91.69 \\
		\Checkmark & - & \Checkmark & 90.99 \\
		- & \Checkmark & \Checkmark & 88.82 \\
		\Checkmark & \Checkmark & \Checkmark & \textbf{91.89} \\
	\bottomrule
	\end{tabular}
\end{adjustbox}
\caption{Ablations of PointCLIP fine-tuning different modules. \Checkmark denotes fine-tuning the module and symbol - denotes freezing.}
\vspace{-0.5cm}
\label{fine-tune}
\end{table}
\section{Visualizations}
We visualize some cases of ensembling PointCLIP with PointNet++~\cite{qi2017pointnet++} to reveal the effectiveness of enhancement. As shown in Figure~\ref{fig:ensemble}, two models both predict correctly for the four samples, and the ensembled model preserves the prediction. As for samples in the second and the third rows, PointCLIP and PointNet++ show the complementary properties that the ensembled model would rectify one of their wrong predictions, which demonstrates the importance of knowledge interaction.

\end{document}